\documentclass[lettersize,journal]{IEEEtran}
\usepackage{bbding}
\usepackage{amsmath,amsfonts}
\usepackage{algorithmic}
\usepackage{array}
\usepackage[caption=false,font=normalsize,labelfont=sf,textfont=sf]{subfig}
\usepackage{textcomp}
\usepackage{stfloats}
\usepackage{url}
\usepackage{multirow}
\usepackage{verbatim}
\usepackage{graphicx}
\usepackage{booktabs}
\usepackage[breaklinks]{hyperref}
\usepackage{balance}
\usepackage{xcolor}

\def\BibTeX{{\rm B\kern-.05em{\sc i\kern-.025em b}\kern-.08em
    T\kern-.1667em\lower.7ex\hbox{E}\kern-.125emX}}

\begin{document}
\title{Natural Language Supervision for Low-light Image Enhancement}

\author{\IEEEauthorblockN{
Jiahui Tang,
Kaihua Zhou,
Zhijian Luo and
Yueen Hou\\}
\IEEEauthorblockA{School of Computer, Jiaying University, Meizhou, R. P. China, 514015\\}
}

\markboth{Natural Language Supervision for Low-light Image Enhancement,~Vol.~18, No.~9, August ~2024}%
{Natural Language Supervision for Low-light Image Enhancement}

\maketitle

\begin{abstract}
With the development of deep learning, numerous methods for low-light image enhancement (LLIE) have demonstrated remarkable performance.
Mainstream LLIE methods typically learn an end-to-end mapping based on pairs of low-light and normal-light images.
However, normal-light images under varying illumination conditions serve as reference images, making it difficult to define a ``perfect'' reference image
This leads to the challenge of reconciling metric-oriented and visual-friendly results.
Recently, many cross-modal studies have found that side information from other related modalities can guide visual representation learning.
Based on this, we introduce a Natural Language Supervision (NLS) strategy, which learns feature maps from text corresponding to images, offering a general and flexible interface for describing an image under different illumination.

However, image distributions conditioned on textual descriptions are highly multimodal, which makes training difficult.
To address this issue, we design a Textual Guidance Conditioning Mechanism (TCM) that incorporates the connections between image regions and sentence words, enhancing the ability to capture fine-grained cross-modal cues for images and text.
This strategy not only utilizes a wider range of supervised sources, but also provides a new paradigm for LLIE based on visual and textual feature alignment.
In order to effectively identify and merge features from various levels of image and textual information, we design a Information Fusion Attention (IFA) module to enhance different regions at different levels.
We integrate the proposed TCM and IFA into a Natural Language Supervision network for LLIE, named NaLSuper.
Finally, extensive experiment demonstrate the robustness and superior effectiveness of our proposed NaLSuper.
\end{abstract}


\section{Introduction}
\IEEEPARstart{H}{igh-quality} images are crucial for various advanced computer vision tasks, e.g., object detection\cite{liu2016ssd}, image classification\cite{al2022low} and semantic segmentation\cite{islam2020semantic}, etc. However, images captured in low-light conditions often suffer from issues like low contrast, low brightness, and serious noises.
Therefore, it is practically important to address brightness degradation to facilitate the exploration of sophisticated dark environments.
To improve the quality of these images, numerous low-light image enhancement (LLIE) methods\cite{pizer1990contrast, fu2015probabilistic, fu2016weighted,guo2016lime, lore2017llnet, guo2020zero, jiang2021enlightengan, xu2022snr} have been proposed in recent years.


\begin{figure}
	\centering
	\includegraphics[width=1\linewidth]{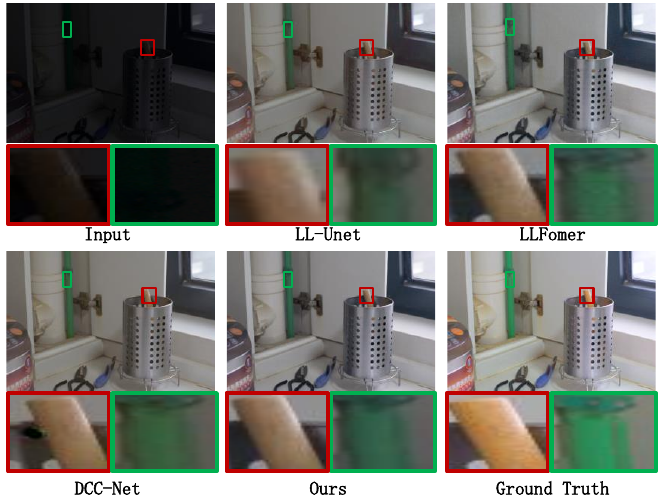}
	\caption{Comparison with state-of-the-art methods on LOLv1 dataset. It is evident that we have restored more authentic colors and visually appealing content.}
	\label{Fig_LOLv1_smoke}
\end{figure}

In recent years, the effectiveness of deep learning methods in computer vision applications has spurred their application in LLIE\cite{ren2019low, guo2020zero} \cite{wei2018deep, lore2017llnet, li2018lightennet, wang2020lightening, xu2020learning, zhao2021retinexdip}.
These deep learning-based methods for LLIE can be categorized into two main groups: Retinex-based methods\cite{wei2018deep, liu2021retinex, ma2021learning}, and end-to-end methods\cite{li2021low,jiang2021enlightengan}.
Retinex-based methods initially decompose the image into illumination and reflectance component using a convolutional neural network.
These components are then processed separately before being recombined to produce the final enhanced image.
However, Retinex-based models are prone to image stylization and noise amplification due to inadequate decomposition of reflectance and illumination, which leads to optimization challenges.
On the other hand, end-to-end methods aim to learn the mapping between low-light and normal-light images without relying on any physical model.
Despite their potential, current end-to-end methods face issues such as underexposure, overexposure, and color imbalance, and their overall generalization performance remains relatively low.

The goal of LLIE is ensuring that the enhanced image's subjective visual experience closely resembles that of a natural image under normal lighting.
To achieve this, many existing methods mainly use normal-light image as a supervisory constraint.
However, the variability in illumination of these normal-light images makes it difficult to define a ``perfect'' reference image.
Furthermore, existing methods primarily focus on generating results that perform well according to specific metrics, often at the expense of visual quality.
Thus these methods face the challenge of reconciling metric-oriented and visual-friendly result.


Recent research in cross-modal learning\cite{radford2021learning, wang2022learning, elizalde2023clap, zhong2021learning} has shown that side information from related modalities can effectively guide visual representation learning.
Based on this, we propose that learning from text corresponding to images is a promising alternative for LLIE, providing a general and flexible interface for describing an image of different illuminations.
Building on this foundation, to enhance low-light image, we propose a Natural Language Supervision strategy, which jointly learns feature maps from text and corresponding image.
However, the highly multimodal distribution of images conditioned on textual descriptions poses training challenges, limiting the application of natural language supervision in LLIE.
The work\cite{wei2020multi} found that incorporating the connections between image regions and sentence words generally, which enhances the ability of network to capture fine-grained cross-modal cues for images and text.
Building upon this inspiration, we introduce a Textual Guidance Conditioning Mechanism (TCM), employing cross-attention to comprehensively capture both cross-modal and intra-modal relationships between image regions and sentence words.


As the network goes deeper and deeper, shallow feature of text and image information is often difficult to preserve.
In order to effectively identify and merge features from various levels of image and textual information, we design a novel information fusion attention (IFA) module to improve feature representation, which can provide additional flexibility in dealing with different modal of information.


In this paper, we propose a Natural Language Supervision network (denoted as NaLSuper) for LLIE, which incorporates TCM and IFA modules.
Overall, our contributions can be summarized as follows:

\begin{itemize}
	\item[1)]
	We propose a Natural Language Supervision network (denoted as NaLSuper) for LLIE, which incorporates Textual Guidance Conditioning Mechanism (TCM) and Information Fusion Attention (IFA) modules.
	\item[2)]
	We are the first to utilize a Natural Language Supervision strategy in LLIE, which use this strategy results in better visual effect of the enhanced image.
	To address the training challenges posed by this strategy in a multi-modal data distribution, we have developed the TCM,  which can comprehensively capture both cross-modal and intra-modal relationships between image regions and sentence words.
	\item[3)]
	We design a novel information fusion attention (IFA) module to improve feature representation, which can provide additional flexibility in dealing with different modal of information.
	Thanks to this module, our network can effectively identify and merge features from various levels of image and textual information.
	 \item[4)]
	Extensive tests conducted on four benchmark dataset reveal that the proposed NaLSuper outperforms recent state-of-the-art methods in both quantitative and qualitative evaluations of quality.
\end{itemize}

\begin{figure*}
	\centering
	\includegraphics[width=1\linewidth]{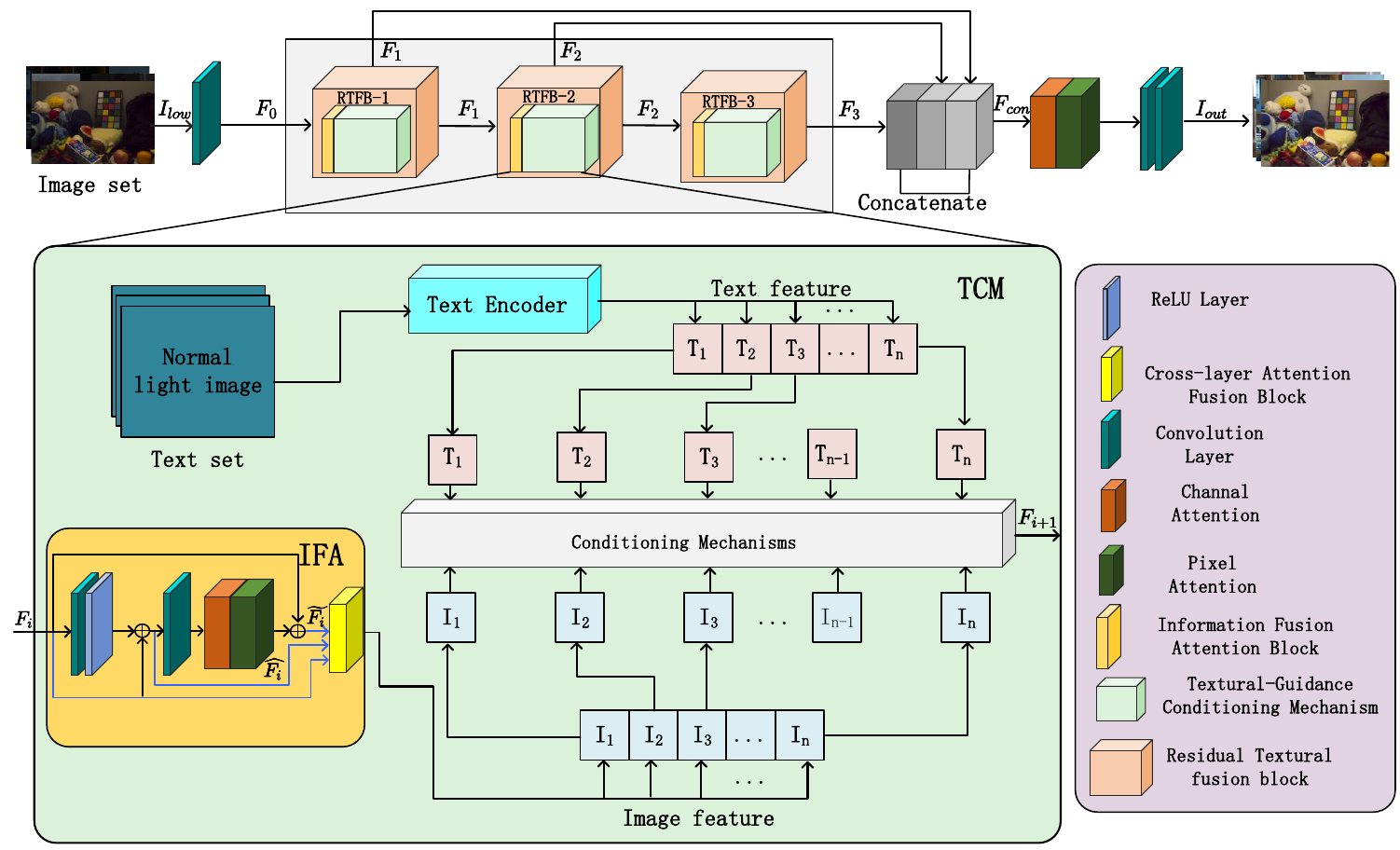}
	\caption{Overview architecture of our proposed NaLSuper. NaLSuper is a Natural Language Supervision network for LLIE, which incorporates Textual Guidance Conditioning Mechanism (TCM) and Information Fusion Attention (IFA) modules. The final estimation outputted by the reconstruction part and global residual learning structure, which is considered to be the desired normal-light image.}
	\label{Fig_architecture}
\end{figure*}

\section{Related Work}
\subsection{Low-light Image Enhancement}
\subsubsection{Traditional Cognition Methods}
In regions with low lighting, pixel values are generally lower.
It is a very forthright idea to directly adjust brightness of image through enhance these lower pixel values, which is referred to as value-based.
The value-based methods include histogram equalization (HE) and gamma correction (GC).
The conventional HE method \cite{cheng2004simple} alters the histogram distribution.
While it adjusts the illumination of the image, it introduces issues such as artifacts, loss of detail, overexposure, and color distortion.
To address these shortcomings, researchers have implemented a series of enhancements to HE.
Kim~\textit{et al.}\cite{kim1997contrast} introduced Brightness Bihistogram Equalization (BBEH), while Wang~\textit{et al.} \cite{wang1999image} proposed Dualistic Subimage Histogram Equalization (DSHE), both of which aimed to produce more natural-looking equalized images.
Nonetheless, the visual artifacts in images equalized by HE methods remain their primary drawback.

GC employs nonlinear transformations to enhance the gray values in darker areas of the image while reducing the gray values in areas with excessively high gray values.
Traditional gamma transformations used for low-light image enhancement suffer from clear drawbacks, including unnatural images, uneven exposure, and loss of details.
To address these issues, Bennett~\textit{et al.}\cite{bennett2005video} enhanced image brightness using Per-Pixel Virtual Exposures.
However, the GC method still suffers from uneven exposure.


\subsubsection{Deep Learning Methods}
With the successful application of deep learning in computer vision \cite{sun2020automatically} and image processing \cite{lan2020madnet}, researchers have turned to deep learning techniques for LLIE.
For instance, Lore~\textit{et al.} \cite{lore2017llnet} introduced LLNet, a stacked sparse denoising auto-encoder designed for simultaneous low-light enhancement and noise reduction.
Wei~\textit{et al.} \cite{wei2018deep} introduced a CNN-based Retinex decomposition method, establishing a deep network that integrates image decomposition and subsequent enhancement operations.
Zhang~\textit{et al.} \cite{zhang2019kindling} merged Retinex theory with convolutional neural networks, dividing the network into two parts: illumination and reflectance estimation, and training it using gamma-corrected simulation data.
Subsequently, Zhang~\textit{et al.} \cite{zhang2021beyond} optimized the model structure based on KinD, resulting in KinD++.
Hao~\textit{et al.} \cite{hao2020low} achieved Retinex image decomposition in a semi-decoupled manner.
Liu~\textit{et al.} \cite{liu2021retinex} proposed a lightweight method called Retinex-inspired Unrolling with Architecture Search (RUAS) for efficient low-light image enhancement.
Ma~\textit{et al.} \cite{ma2021learning} summarized deep learning methods based on Retinex theory and  a context-sensitive decomposition network, along with supervised and self-supervised versions.
Zhu~\textit{et al.} \cite{zhu2022enlightening} proposed a method involving a learnable guidance map from signal and deep priors, enabling adaptive enhancement of low-light images in a region-dependent manner.
Zhao~\textit{et al.} \cite{zhao2021retinexdip} introduced a Retinex decomposition "generative" strategy, which formed the basis for a unified deep framework to estimate latent components and enhance low-light images.
Wu~\textit{et al.} \cite{wu2022uretinex} developed three neural modules that facilitate image recovery in three phases: initialization, optimization, and illumination adjustment.
The previous methods aimed at LLIE were either designed within a single scale framework or implemented through a cascaded process, which limits their effectiveness across various low-light conditions.

\subsection{Natural Language Supervision}
Vision-language pretraining has recently emerged as a promising approach for understanding images \cite{jia2021scaling, radford2021learning, wang2022learning} and videos \cite{wang2021actionclip}.
Unlike traditional approaches that rely on discrete labels, it introduces a new recognition paradigm based on the alignment of visual and textual features.
Recent studies have also explored how to utilize the transferable knowledge of pre-trained models for tasks such as visual question answering (VQA) \cite{jin2021good}, zero-shot object detection \cite{gu2021open}, and image captioning \cite{shen2021much}.
For example, Radford et al. \cite{radford2021learning} introduced CLIP, a method for learning visual models under language supervision. After being trained on 400 million image-text pairs, CLIP is capable of describing any visual concept in natural language and can be applied to other tasks without further specific training.
Zhou~\textit{et al.}. \cite{zhou2022learning} introduced soft prompts, utilizing learnable vectors to model context words instead of traditional hand-crafted prompts, thereby capturing task-relevant context.
Rao~\textit{et al.} \cite{rao2022denseclip} innovatively introduced context-aware prompting, integrating prompts with visual features for more precise instance-level refinement.
Cho~\textit{et al.} \cite{cho2021unifying} harmonized prior knowledge across various tasks through a unified framework tailored to seven multi-modal tasks.
Additionally, Ju~\textit{et al.} \cite{ju2022prompting} leveraged the pre-trained CLIP model to enhance video comprehension.
In this paper, we aim to explore how natural language supervision can be utilized for LLIE.

\section{Method}
In this section, we mainly introduce a Natural Language Supervision network (denoted as NaLSuper) for LLIE.
As illustrated in Fig. \ref{Fig_architecture}, the input of NaLSuper is a low light image $I_{low} \in \mathbb{R}^{H\times W \times 3}$, it is first passed into a  $3\times3$ convolution as a projection layer to extract shallow feature $F_0\in \mathbb{R}^{H\times W \times C}$.
Next, $F_0$ is fed into three Residual Textual guide Fusion Block (RTFB) with multiple skip connections, which can extract deeper feature of fine-grained cross-modal information for images and text.
Specifically, intermediate features outputted from RTFB are denoted as $F_1, F_2, F_3 \in \mathbb{R}^{H\times W \times C}$.
After that, these features $F_1, F_2, F_3$ will be concatenate to $F_{con} \in \mathbb{R}^{H\times W \times 3C}$, and then it passed to the reconstruction part and global residual learning structure, thereby getting a enhanced image $I_{out}\in \mathbb{R}^{H\times W \times 3}$.

Furthermore, we combines RTFB Architecture with local residual learning, every RTFB combines the Textual Guidance Conditioning Mechanism (TCM) and Information Fusion Attention (IFA) modules.

\subsection{Textual Guidance Conditioning Mechanism (TCM)}
Current approaches predominantly rely on image-level supervision, where the output is constrained to closely resemble target images.
Yet, these methods faces challenges due to significant brightness discrepancies among different references, which can complicate model training.
Additionally, some references exhibit visual flaws, such as unnatural brightness, resulting in outputs that are visually unappealing.
To alleviate training complexities and reconcile the disparity between metric-oriented and visually pleasing versions, as shown in Fig. \ref{Fig_architecture}, we design a Textual Guidance Conditioning Mechanism (TCM) to comprehensively capture both cross-modal and intra-modal relationships between image regions and sentence words.
\subsubsection{Text Encoder}
The main objective of the text encoder is to map the raw textual descriptions of interactions to the feature space.
The raw text will be broken down into tokens and transformed into a series of word embeddings.
Recent research \cite{radford2021learning, zhou2022learning} indicates that the selection of context words around the class name can greatly affect the accuracy of recognition.
In this work, we employ the recent well-known pretrained CLIP\cite{radford2021learning} text encoder and keep it fixed during training.
As shown in Fig. \ref{Fig_architecture}, We first manually design a series of prompt, such as normal light image.
After CLIP, text feature $T_i, [i=1, 2,\dots, n]$ is obtained, which has textual semantic information of corresponding image.
%
\subsubsection{Conditioning Mechanisms}
Image distributions conditioned on textual descriptions are highly multimodal, which makes training difficult.
To address this problem, we use the cross-attention mechanism as the fusion layer (refer to Fig. \ref{Fig_architecture}), which is valid for attention-based models that learn relationships between various input modalities \cite{jaegle2021perceiver}.
To preprocess $T_i$ from different modalities (such as language prompts), we utilize a domain-specific encoder $\tau_{\theta}$ that maps $T_i$ to an intermediate representation $\tau_{\theta}(T_i)$ in $\mathbb{R}^{d_{\tau} \times M}$.
Then the computation of query matrix $Q$, key matrix $K$, and value matrix $V$ are as follow:
\begin{equation}
	Q=W_q \cdot  \psi (I_i),K=W_k \cdot \tau_{\theta}(T_i),V=W_v \cdot \tau_{\theta}(T_i),
\end{equation}
where $W_q  \in \mathbb{R}^{d \times d_{\zeta}}$, $W_k \in \mathbb{R}^{d \times d_{\tau}}$, and $W_v \in \mathbb{R}^{d \times d_{\tau}}$ represent the projection matrices\cite{jaegle2021perceiver, vaswani2017attention}, $I_i, [i=1, 2,\dots, n]$ represent a feature map of image and $\psi (I_i)\in \mathbb{R}^{d_{\zeta} \times M}$ denotes a (flattened) intermediate representation.
Suppose that we have $Q,K,V \in  \mathbb{R}^{d \times M}$, and the attention matrix is expressed as:
\begin{equation}
		\text{Attention}(Q,K,V)=\text{softmax} \left(\frac{Q K^T}{\sqrt{d} } +B \right)V,
	\label{eq:coss-attention}
\end{equation}
where $B$ is a learnable matrix of relative position encoding. Refer to Fig. \ref{Fig_architecture} for a visual representation.

\begin{figure}
	\centering
	\includegraphics[width=1\linewidth]{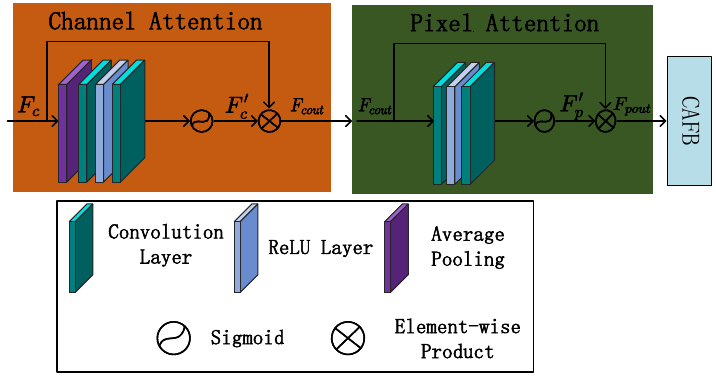}
	\caption{Overview architecture of Information Fusion Attention (IFA).}
	\label{Fig_IFA}
\end{figure}

\subsection{Information Fusion Attention (IFA)}

Many LLIE networks uniformly process channel-wise and pixel-wise features, which fails to effectively address images with uneven illumination distribution and weighted channel-wise features.
Additionally, the network goes deeper and deeper, shallow feature of text and image information is often difficult to preserve.
In order to effectively identify and merge features from various levels of image and textual information.
We design a Information Fusion Attention (IFA) module (refer to Fig. \ref{Fig_IFA}) includes both channel attention, pixel attention and Cross-layer Attention Fusion Block, which can provide additional flexibility in dealing with different modal of information.

\subsubsection{Channel Attention}
Our channel attention focuses on the concept that various channel features carry distinctly different weights of information in relation to DCP \cite{he2010single}.
Inspired by \cite{qin2020ffa}, we capture the channel-wise global spatial information through a channel descriptor by employing global average pooling.
\begin{equation}
	g_c = P(F_c) = \frac{1}{H \times W} \sum_{i=1}^{H} \sum_{j=1}^{W} X_c(i, j)
	\label{eq:CA_pooling}
\end{equation}

Eq. \ref{eq:CA_pooling} represents the global average pooling operation $P(\cdot)$ applied to a feature map. $g_c$ is the pooled feature for the channel $c$, obtained by averaging all pixel values $X_c(i,j)$ across the spatial dimensions $H$ (height) and $W$ (width) of the feature map $F_c$.
This reduces the spatial dimensions of each channel to $1 \times 1$, effectively compressing the spatial information of the feature map into a single value per channel.

To determine the weights of various channels, the features go through two convolutional layers and subsequently are activated using sigmoid and ReLU functions.
Finally, an element-wise multiplication is performed between the input feature map $F_c$ and the channel-specific weights . The Channel Attention process is formulated as:
\begin{equation} \label{eq:CA}
	\begin{split}
		& F'_c = \text{Sigmoid}(\text{Conv}(\text{ReLu}(\text{Conv}(g_c)))), \\
		& F_{cout}=CA(F_c) =  F'_c \otimes F_c. 	
	\end{split}
\end{equation}

\begin{figure*}
	\centering
	\includegraphics[width=1\linewidth]{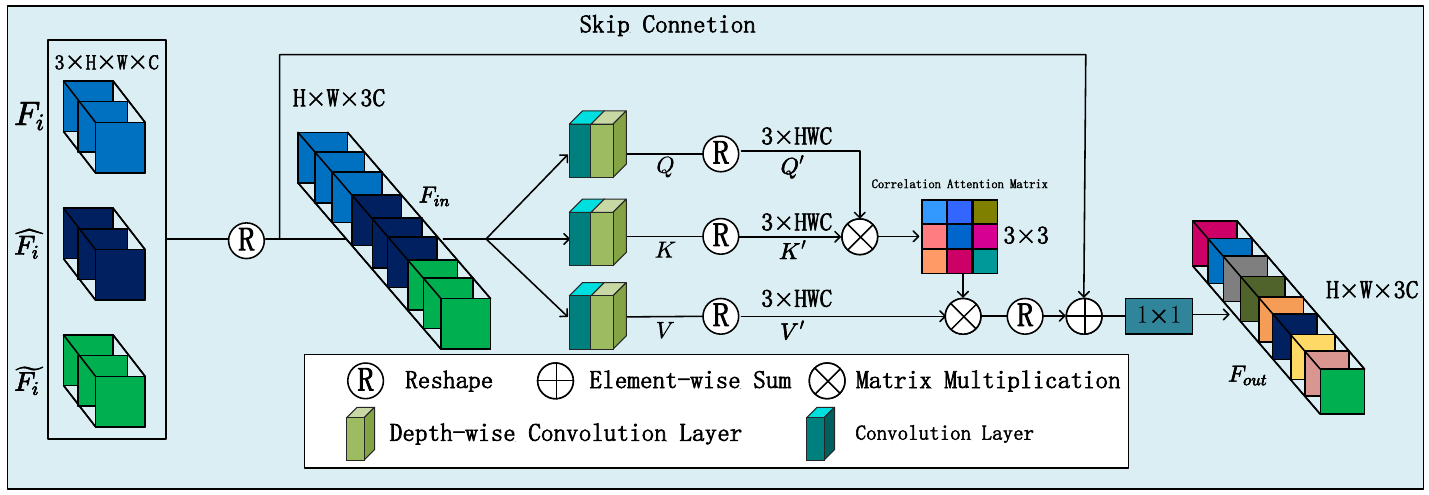}
	\caption{Overview architecture of Cross-layer Attention Fusion Block(CAFB).}
	\label{Fig_CAFB}
\end{figure*}

\subsubsection{Pixel Attention}
Given the uneven distribution of illumination across different pixels in an image, we employ a pixel attention module to make the network pay more attention to informative features. This includes areas with low-light and regions of the image with high frequency.

Similar to \cite{qin2020ffa}, we directly feed the input $F_{cout}$ into two convolutional layers and subsequently are activated using sigmoid and ReLU functions.
Finally, an element-wise multiplication is performed between the input feature map $F_{cout}$ and the pixel-specific weights. The Pixel Attention process is formulated as:
\begin{equation} \label{eq:PA}
	\begin{split}
		& F'_p = \text{Sigmoid}(\text{Conv}(\text{ReLu}(\text{Conv}(F_{cout})))),  \\
		& F_{pout} = CA(F_c) = F'_p \otimes F_{cout}.
	\end{split}
\end{equation}

\subsubsection{Cross-layer Attention Fusion Block}
Recent methods using transformers incorporate feature connections or skip connections to merge features across various layers, as discussed in studies by Zamir et al. \cite{zamir2022restormer} and Wang et al. \cite{wang2022uformer}. Nevertheless, these methods do not completely leverage the inter-layer dependencies, which restricts their ability to represent complex features effectively.
To address this problem, we use a Cross-layer Attention Fusion Block(CAFB) \cite{wang2023ultra}, which adaptively fuses hierarchical features with learnable correlations among different layers.
The underlying concept of CAFB is that activations in various layers correspond to specific classes, and the correlations between these features can be dynamically learned through a self-attention mechanism.

As show in Fig. \ref{Fig_architecture}, given features $F_{i}, \widehat{F_i}, \widetilde{F_i}  \in \mathbb{R}^{H\times W \times C}$, we concatenate and reshape them to make $F_{in} \in \mathbb{R}^{H\times W \times 3C}$.
Following \cite{wang2023ultra}, we use 1$\times$1 convolutional layers to integrate context across different channels at the pixel level, and then apply 3$\times$3 depth-wise convolutions to generate $Q$, $K$, and $V$.
We reshape the matrices $Q$ and $K$ into 2D forms with dimensions $3 \times HWC$ (denoted as $Q'$) and $3 \times HWC$ (denoted as $K'$), respectively.
This restructuring allows us to compute the layer correlation attention matrix $A$, which has a size of $3 \times 3$.
Finally, we scale the reshaped value $V' \in \mathbb{R}^{HWC \times 3}$ by multiplying it with the attention matrix $A$ and a scaling factor $\delta $.
\begin{equation} \label{eq:coss_layer}
	 \text{Cross-Layer}(Q', K', V') = \text{softmax} \left(\frac{Q' K'}{\delta} \right)V',
\end{equation}
Then, we add the input features $F_{in}$ to this product.
The CAFB process is formulated as:
\begin{equation} \label{eq:coss_layer_block}
	F_{out} = \text{Cross-Layer}(Q', K', V') + F_{in} ,
\end{equation}
$F_{out}$ represents the output feature that targets the informative layers within the network.
In practical terms, we strategically position the CAFB at locations in the tail of each RTFB.
This placement enables CAFB to effectively capture long-distance dependencies across hierarchical layers during both the feature extraction and image reconstruction phases.

\subsection{Loss Function}
There are two reconstruction loss terms to train our framework, i.e., the $L_1$ loss and the SSIM loss. The L1 loss is written as
\begin{equation} \label{eq:L1}
	L_1 = \frac{1}{n}\sum_{i=1}^{n}|| I_{gt}^i - \text{enhanced}(I_{low}^i)||,
\end{equation}
where $I_{gt}$ denotes the ground truth, $n$ is the batch size and $I_{low}$ stands for input(low-light image).
The SSIM loss is written as
\begin{equation} \label{eq:SSIM}
		 L_{SSIM} = \frac{1}{n}\sum_{i=1}^{n}(\frac{(2 \mu_{I_{gt}^i}\cdot \mu_{I_{low}^i} + c_1) \cdot (2 \sigma_{I_{gt}^i,I_{low}^i} + c_2)}{(\mu_{I_{gt}^i}^2 + \mu_{I_{low}^i}^2 + c_1) \cdot (\sigma_{I_{gt}^i}^2 + \sigma_{I_{low}^i}^2 + c_2)}),
\end{equation}
where $\mu$ is the variances, and $\sigma$ is the covariance. $c_1$ and $c_2$ are constants added to avoid having a denominator of zero.
The overall loss function is
\begin{equation} \label{eq:total_loss}
	L = L_1 + L_{SSIM}.
\end{equation}

\begin{figure*}
	\centering
	\includegraphics[width=1\linewidth]{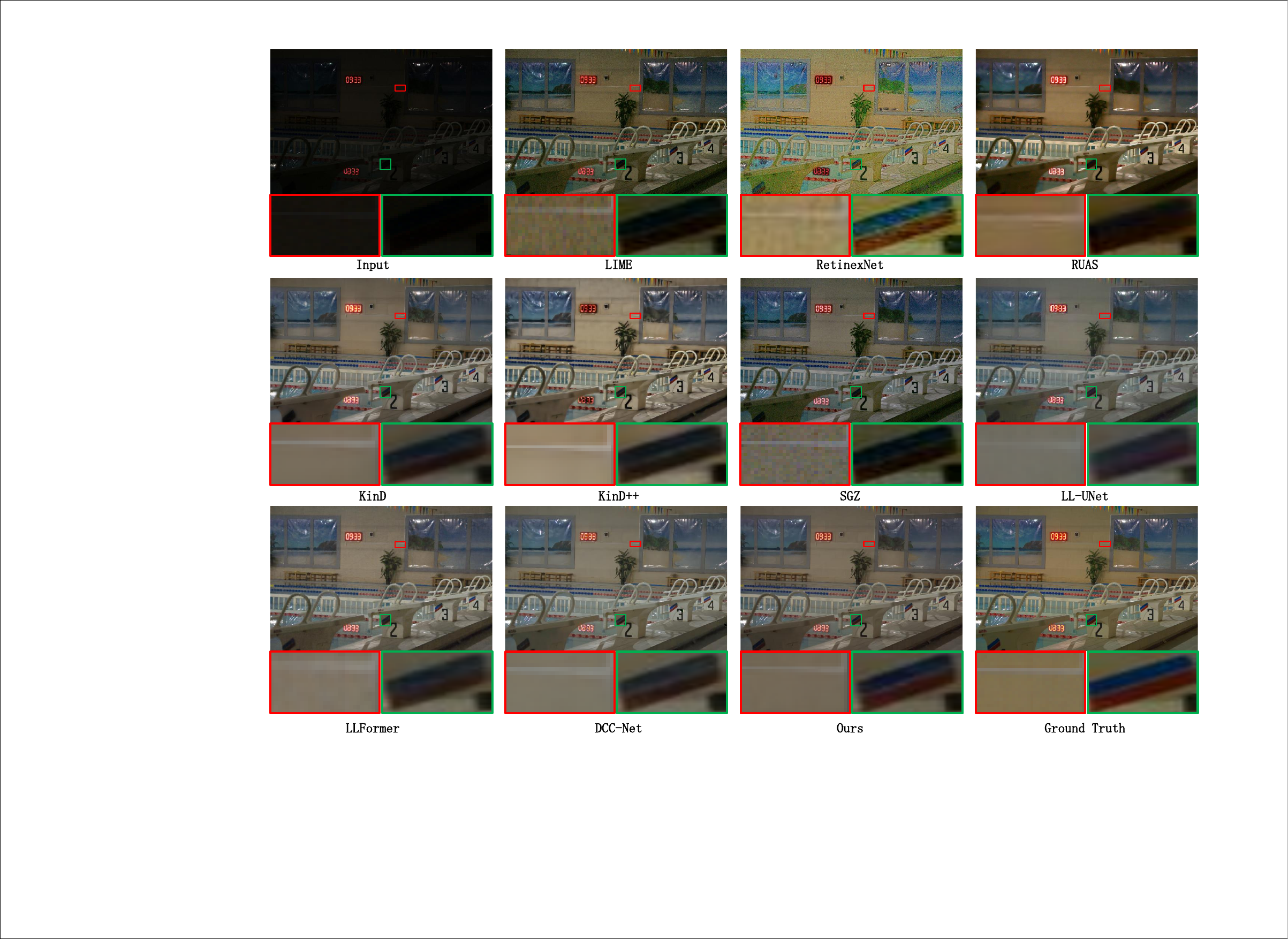}
	\caption{Visual comparison with LLIE methods on LOLv1 dataset.}
	\label{Fig_LOLv1_pooling}
\end{figure*}

\section{Experimental Results}
\subsection{Benchmark Datasets }
To evolution the performance and efficiency of the proposed NaLSuper, we test our approach on several publicly accessible low-light datasets, such as LOLv1 \cite{wei2018deep},  LOLv2, \cite{yang2020fidelity} and SID \cite{chen2018learning}.

\begin{figure*}
	\centering
	\includegraphics[width=1\linewidth]{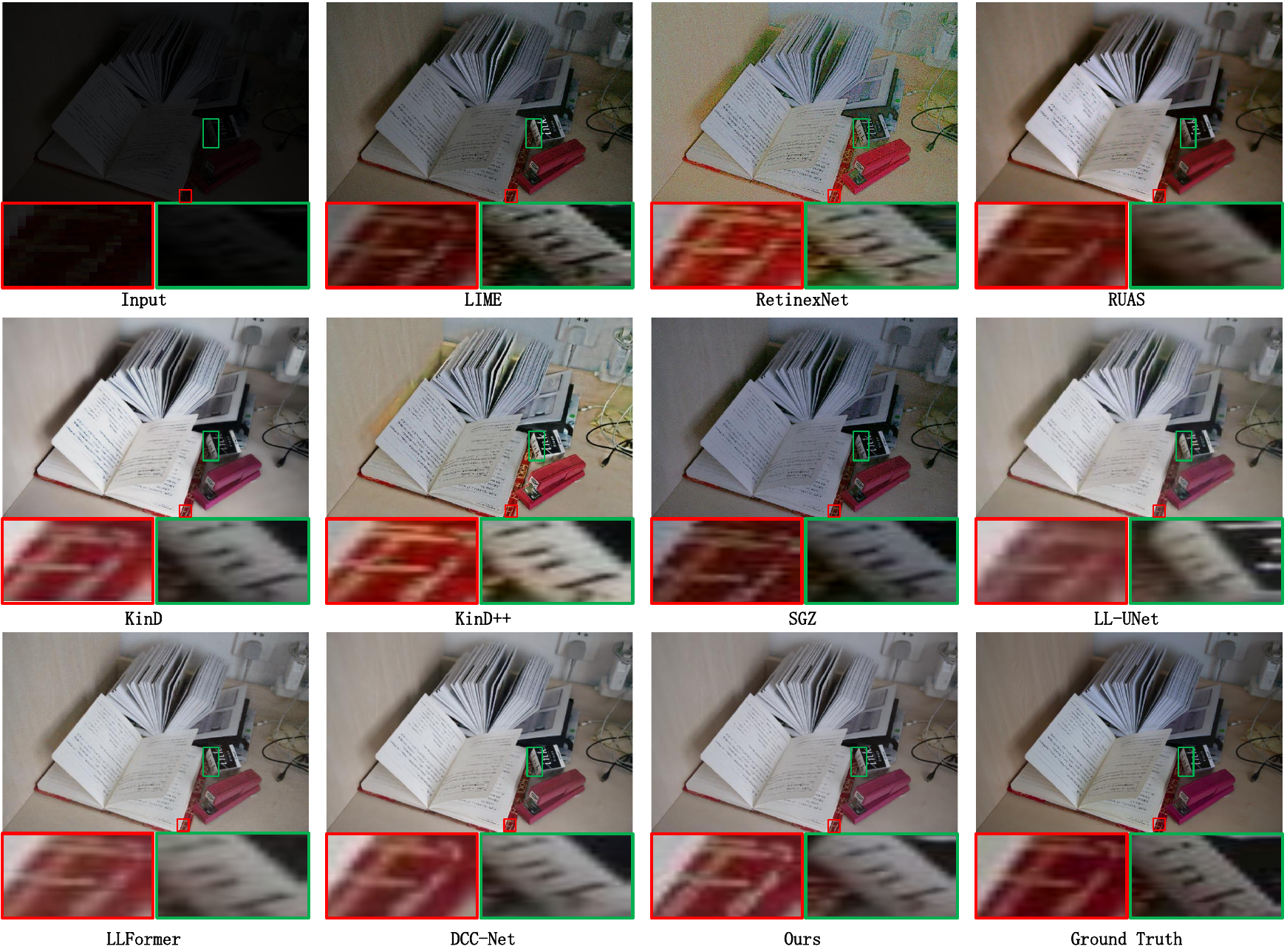}
	\caption{Visual comparison with LLIE methods on LOLv1 dataset.}
	\label{Fig_LOLv1_book}
\end{figure*}

The LOL dataset is available in two versions: v1 and v2.
LOLv1 provides 485 pairs of low-/normal-light images for training and 15 pairs for testing.
Each pair consists of a low-light image and its corresponding well-exposed reference image.
LOLv2 is split into two subsets: LOLv2-real and LOLv2-synthetic.
The training set for LOLv2-real includes 689 pairs of low-/normal-light images, and the test set comprises 100 pairs.
These low-light images are primarily captured in various settings by adjusting the ISO and exposure time, while keeping other parameters constant.
LOL-v2-synthetic, however, is generated by analyzing the light distribution in low-light images and then synthesizing these from RAW images.

The subset of the SID dataset captured with the Sony $\alpha 7S$ II camera is used for evaluation. It comprises 2697 pairs of short/long-exposure RAW images. The low-/normal-light RGB images are derived by applying the same in-camera signal processing used in SID to convert the RAW images into the RGB format. Specifically, 2099 image pairs are utilized for training, while 598 pairs are designated for testing.


The four widely adopted quality metrics, i.e., peak signal-to-noise ratio (PSNR), structural similarity (SSIM)\cite{wang2004image}, LPIPS and mean absolute error (MAE), are adopted for quantitative comparison as evaluation metrics between the enhanced images and the referenced ground truths. For the metrics PSNR and SSIM, higher values signify improved image quality. Conversely, a lower LPIPS and MAE score indicates superior quality.


\subsection{Implementation Details}
For model training of our proposed network, we use the Adam optimizer with a learning rate of $10^{-4}$.
All experiments on benchmark datasets are implemented with PyTorch, on a 64 core Intel Xeon Gold 6226R CPU @2.90GHz, 256 GB memory and a Nvidia Duadro RTX 8000 GPU.

\begin{table}
	\begin{center}
		\scalebox{1}{\begin{tabular}{|l|c|c|c|c|}
				\hline
				\multirow{2}{*}{Methods}  & \multicolumn{4}{c|}{\textbf{LOLv1}}  \\ \cline{2-5}
				
				& PSNR $\uparrow$ & SSIM $\uparrow$ & LPIPS $\downarrow$ & MAE $\downarrow$    \\ \hline

				BIMEF \cite{ying2017bio}  & 13.88 & 0.595 & 0.3264 & 0.2063  \\ \cline{2-5}
				FEA \cite{dong2010fast}  & 16.72 & 0.478 & 0.3847 & 0.1421  \\ \cline{2-5}
				LIME \cite{guo2016lime} & 16.76 & 0.445 & 0.3945 & 0.1200    \\ \cline{2-5}
				MF \cite{fu2016fusion}  & 16.97 & 0.508 &  0.3796  & 0.1416  \\ \cline{2-5}
				NPE \cite{wang2013naturalness} & 16.97 &  0.484  &  0.4049 & 0.1290  \\ \cline{2-5}
				
				SRIE \cite{fu2016weighted}  & 11.86 & 0.495 & 0.3401 & 0.2571  \\  \cline{2-5}
				
				MSRCR \cite{jobson1997multiscale} &  13.17 & 0.462 & 0.4350 & 0.2067 \\ \cline{2-5}
				
				RetinexNet \cite{wei2018deep}   & 16.77 & 0.425 & 0.4739& 0.1256   \\ \cline{2-5}
				
				DSLR \cite{lim2020dslr}   & 14.98 & 0.596  & 0.3757&0.1918      \\ \cline{2-5}
				
				KinD \cite{zhang2019kindling}   & 17.65 & 0.772 & 0.1750 & 0.1231 \\ \cline{2-5}
				
				Z\_DCE \cite{guo2020zero} &  14.86 & 0.562 & 0.3352 &0.1846  \\ \cline{2-5}
				
				Z\_DCE++ \cite{li2021learning} &  14.75 & 0.518  & 0.3284  & 0.1801  \\ \cline{2-5}
				MIRNet \cite{zamir2020learning}  & 20.34 & 0.786 &0.1688 & 0.1123   \\ \cline{2-5}
				
				RUAS \cite{liu2021retinex} &   16.40  &  0.503 & 0.2701  &  0.1534     \\ \cline{2-5}
				
				ELGAN \cite{jiang2021enlightengan} & 17.48 & 0.652  & 0.3223  & 0.1352     \\ \cline{2-5}
				
				Uformer \cite{wang2022uformer}  & 18.55 & 0.721 & 0.3205 & 0.113
				\\ \cline{2-5}
				
				Restormer \cite{zamir2022restormer} & 22.37 & 0.816 &0.1413& 0.0721
				\\ \cline{2-5}


				LL-Unet\cite{shi2024ll} &  21.46 & 0.789 & 0.2077  &  0.2630
				\\ \cline{2-5}

				DDC-net \cite{zhang2022deep} &  22.98 & \textcolor{blue}{0.851} &  \textcolor{blue}{0.0905} & 0.0856
				\\ \cline{2-5}

				LLFormer \cite{wang2023ultra} &  \textcolor{blue}{23.65} & 0.816 &  0.1692& \textcolor{red}{0.0754}
				\\ \cline{2-5}

				\textbf{NalSper} &  \textcolor{red}{24.01} & \textcolor{red}{0.863} &  \textcolor{red}{0.0747}& \textcolor{blue}{0.0756}
				\\ \hline
		\end{tabular}}
		\vspace{0.1in}
		\caption{Comparison of average PSNR/SSIM/LPIPS/MAE on LOLv1 test dataset. The best results are marked in  \textcolor{red}{red} color and the second best results are marked in \textcolor{blue}{blue} color.}
		\label{tab:LOLv1}
	\end{center}
\end{table}

\subsection{Quantitative Evaluation}

\begin{figure*}
	\centering
	\includegraphics[width=1\linewidth]{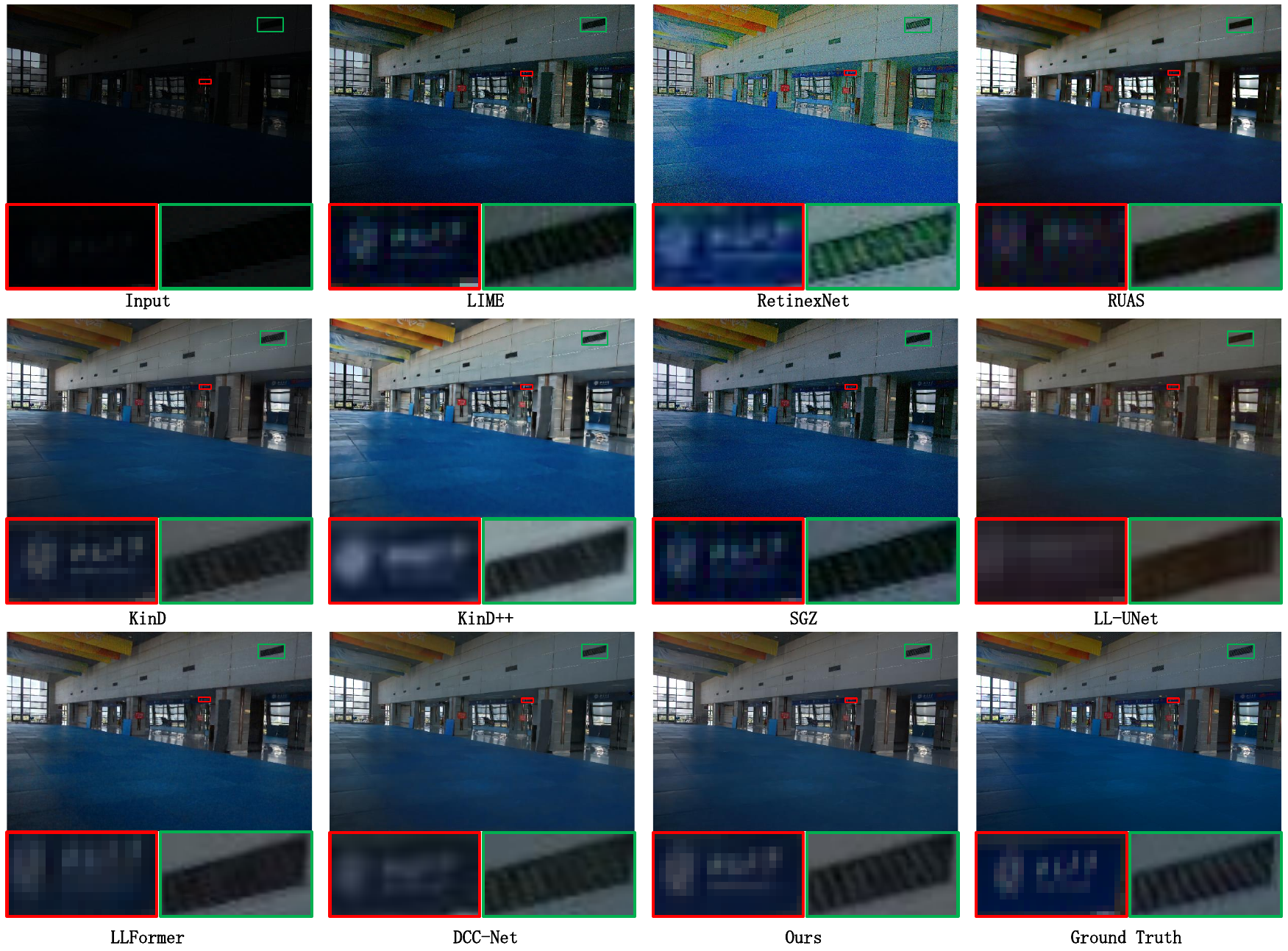}
	\caption{Visual comparison with LLIE methods on LOLv1 dataset.}
	\label{Fig_LOLv1_room}
\end{figure*}

\begin{table*}[t]\small
	\begin{center}
		\vspace{-0.1in}
		\scalebox{0.9}{\begin{tabular}{|l|c|c|c|c|c|c|}
				\hline
				\multirow{2}{*}{Methods}  & \multicolumn{2}{c|}{\textbf{LOLv2-real}}   & \multicolumn{2}{c|}{\textbf{LOLv2-syn}}  & \multicolumn{2}{c|}{\textbf{SID}}   \\ \cline{2-7}
				
				& PSNR $\uparrow$ & SSIM $\uparrow$  & PSNR $\uparrow$ & SSIM $\uparrow$  & PSNR $\uparrow$ & SSIM $\uparrow$ \\ \hline
				KinD \cite{zhang2019kindling}   & 17.54 & 0.669 & 13.29 & 0.578 & 18.02 & 0.583 \\ \cline{2-7}
				
				RetinexNet \cite{wei2018deep}   & 16.10 & 0.410 & 17.13 & 0.809 & 16.48 & 0.578   \\ \cline{2-7}
				ELGAN \cite{jiang2021enlightengan} & 18.64 & 0.670  & 16.57 & 0.737 &17.23 & 0.543   \\ \cline{2-7}
				RUAS \cite{liu2021retinex} &18.37 & 0.723  & 16.55 & 0.652 &18.44 &0.581   \\ \cline{2-7}
				
				UFormer \cite{wang2022uformer} & 18.82 & 0.771 & 19.66 &0.871 & 18.54 &0.577   \\ \cline{2-7} 		
				MIRNet \cite{zamir2020learning} & 20.02 & 0.820 &21.94 & 0.876 & 20.84 & 0.605   \\ \cline{2-7}
				
				Restormer \cite{zamir2022restormer}   & 19.94 & \textcolor{blue}{0.827} & 21.41 & 0.830 & 22.27 & 0.649 \\ \cline{2-7}
				LLformer \cite{wang2023ultra} & \textcolor{red}{21.73} & 0.819 &\textcolor{blue}{23.24} & \textcolor{blue}{0.894} & -- & --   \\ \cline{2-7}
				\textbf{NalSper} &  \textcolor{blue}{21.12} & \textcolor{red}{0.840} &  \textcolor{red}{24.48} &  \textcolor{red}{0.929} & \textcolor{red}{22.22} &  \textcolor{red}{0.610}
				\\ \hline
		\end{tabular}}
		\vspace{0.1in}
		\caption{Comparison of average PSNR/SSIM on LOLv2-real, LOLv2-syn and SID test dataset. The best results are marked in  \textcolor{red}{red} color and the second best results are marked in \textcolor{blue}{blue} color.}
		\label{tab:results2}
		\vspace{-0.15in}
	\end{center}
\end{table*}

To evaluate the advantages of proposed NaLSuper, we quantitatively compare our proposed method with several LLIE state-of-the art competitors, including BIMEF \cite{ying2017bio}, FEA \cite{dong2010fast}, LIME \cite{guo2016lime}, MF \cite{fu2016fusion}, NPE \cite{wang2013naturalness}, SRIE \cite{fu2016weighted}, MSRCR \cite{jobson1997multiscale}, RetinexNet \cite{wei2018deep}, DSLR \cite{lim2020dslr}, KinD \cite{zhang2019kindling}, MIRNet \cite{zamir2020learning}, Z\_DCE \cite{guo2020zero}, Z\_DCE++ \cite{li2021learning}, RUAS \cite{liu2021retinex}, ELGAN \cite{jiang2021enlightengan}, Uformer \cite{wang2022uformer}, Restormer \cite{zamir2022restormer}, LL-Unet \cite{shi2024ll}, DDC-net \cite{zhang2022deep}, LLFormer \cite{wang2023ultra}.
We utilize the publicly available source code and recommended parameters for each of the compared methods, and fine-tune all models on the training dataset to enable comparison.

Table \ref{tab:LOLv1} tabulates the comparisons of averaged PSNR/SSIM/LPIPS/MAE scores tested on the LOLv1 testset.
The top two results are marked in \textcolor{red}{red}, \textcolor{blue}{blue}.
The comparative results presented in the table clearly demonstrate that the proposed NaLSuper significantly outperforms other state-of-the-art methods on the LOLv1 testset.
More specifically, our approach achieves the highest PSNR score compared to all other methods and exhibits substantial improvements, with gains reaching up to 0.36dB over the second-best results for the LOLv1 testset.
The outstanding performance suggests that our method excels in producing results, wihch are highly faithful to the ground truths.
Additionally, our approach also achieves the best SSIM and LPIPS scores compared to other competitors.
Unlike PSNR, the SSIM and LPIPS quality metrics align more closely with the human visual system (HVS), providing a better consistency.
As shown in Table \ref{tab:LOLv1}, the SSIM improvements reach up to 0.012 above the second-best results, while the LPIPS improvements reduce to 0.0158 below the second-best results for the LOLv1 test set.
Moreover, the MAE scores also achieves the secondly compared to other competitors.

Table \ref{tab:results2} tabulates the comparisons of averaged PSNR/SSIM scores tested on the LOLv2-real, LOL-v2-syn and SID testset.
The comparative results presented in the table clearly demonstrate that the proposed NaLSuper significantly outperforms other state-of-the-art methods on the LOLv2-real, LOLv2-syn and SID testset.
More specifically, our approach achieves the highest PSNR score compared to all other methods and exhibits substantial improvements, with gains reaching up to ---dB, ---dB and ---dB  over the second-best results for the LOLv2-real, LOL-v2-syn and SID testset.
The outstanding performance suggests that our method excels in producing results, wihch are highly faithful to the ground truths.
Additionally, our approach also achieves the best SSIM scores compared to other competitors.
As shown in Table \ref{tab:results2}, the SSIM improvements reach up to --- above the second-best results for the LOLv2-real, LOL-v2-syn and SID testset.
The remarkable results confirm that our method not only delivers the most accurate reconstruction but also excels in performance according to perceptual-oriented assessment outcomes.

\begin{figure*}
	\centering
	\includegraphics[width=1\linewidth]{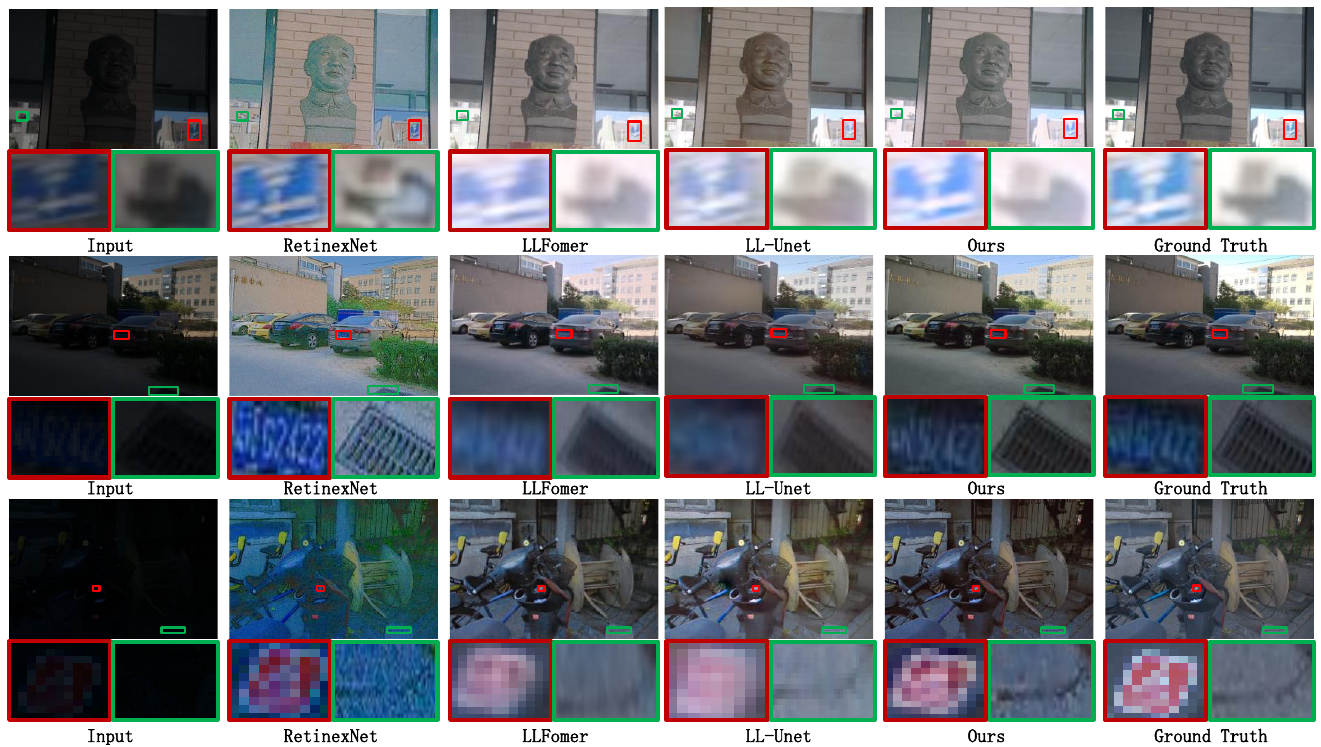}
	\caption{Visual comparison with LLIE methods on LOLv2-real dataset.}
	\label{Fig_LOLv2_statue}
\end{figure*}

\subsection{Subjective Evaluation}
To further demonstrate the superiority of the proposed low-light image enhancement method over other competitors, we have selected nine methods with higher PSNR values for comparison on the LOLv1 and LOLv2-real dataset.
The visual quality comparison results are shown in Fig. \ref{Fig_LOLv1_pooling}--\ref{Fig_LOLv2_statue}.
The images contained in the LOLv1 and LOLv2-real dataset are taken indoors and are very dark and noisy, which is very challenging for many LLIE methods.
The images with the LOLv1 and LOLv2-real dataset are captured indoors, characterized by low light and high noise levels, presenting a significant challenge for numerous LLIE methods.
Based on the comparison results, our method demonstrates the most realistic outcomes, exhibiting minimal artifacts and noise in the predominant regions.
Moreover, the enhanced results exhibit the most faithful representation of the target image in terms of brightness, contrast, and color.
In comparison, other methods that were evaluated tend to create lots of unexpected halos in edge and textural areas, and some even cause noticeable color distortions across the reconstructed images.
The proposed method benefits from the carefully designed natural language supervision strategy, enabling it to effectively address multiple complex degradations in extremely dark environments.
It demonstrates proficient restoration of brightness, color, contrast, and detail.

\begin{figure*}
	\centering
	\includegraphics[width=1\linewidth]{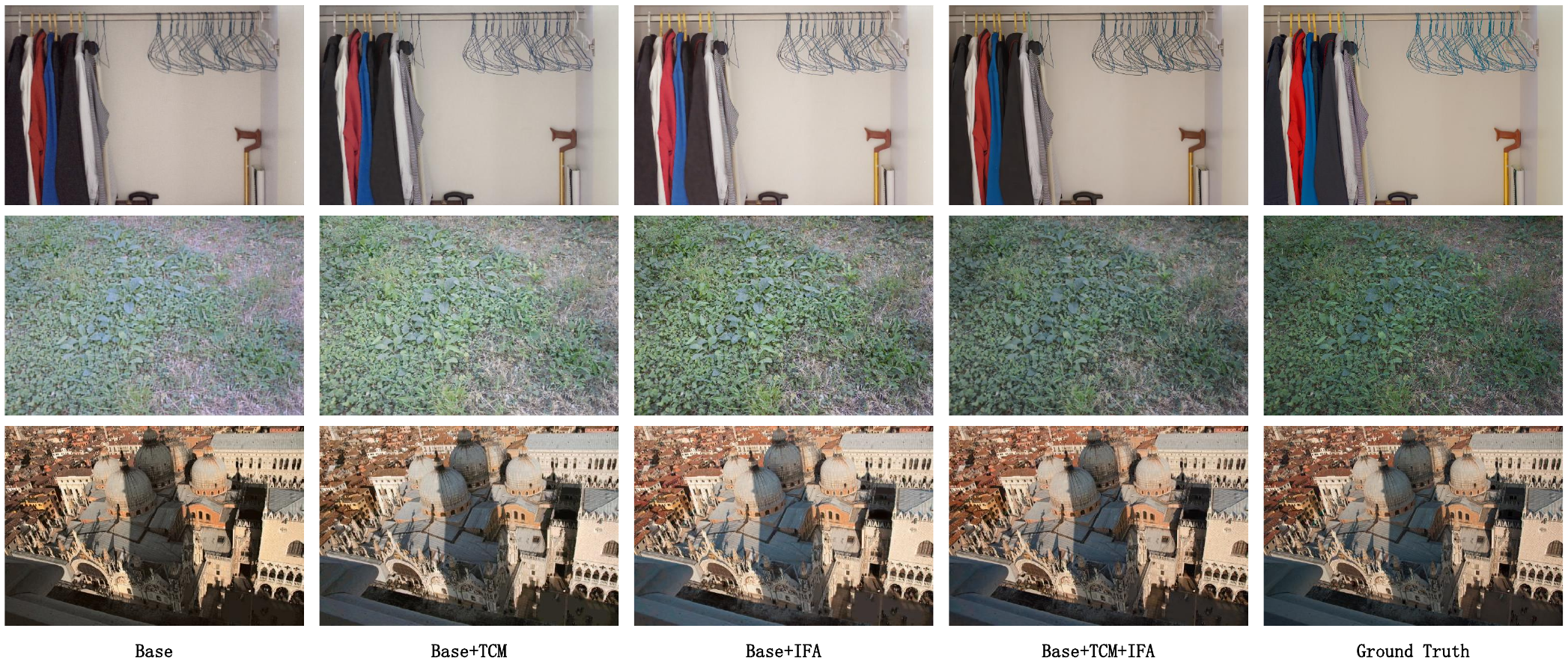}
	\caption{Comparison of the subjective visual effects of models with varying configurations.}
	\label{ablation_module_image}
\end{figure*}

\begin{figure}
	\centering
	\includegraphics[width=1\linewidth]{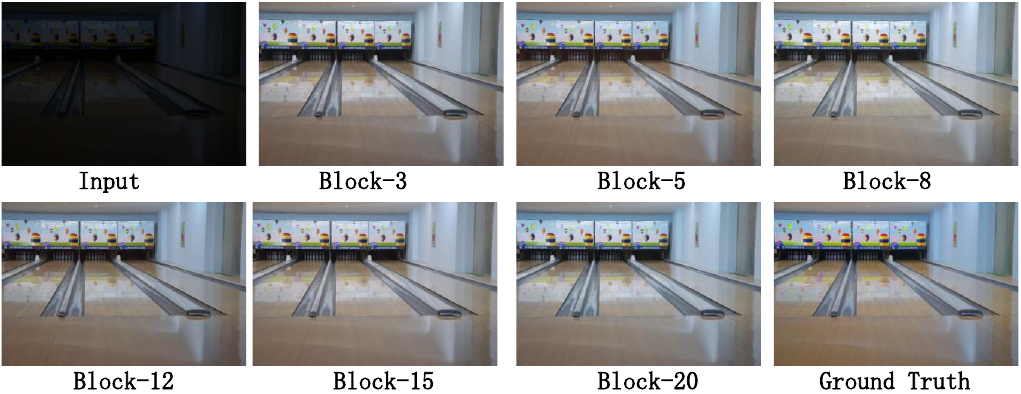}
	\caption{Comparison of the subjective visual effects of models with different numbers of RTFB .}
	\label{ablation_block_image}
\end{figure}

\subsection{Ablation Study} \label{sec:ablation}
\subsubsection{Model Architecture}

To verify the advantages of the network architecture, we conducted an ablation study.
We refer to the simplest version of the network, which lacks any specialized modules, as "base".
Building on this "base" model, we introduce various configurations to determine if the ultimate model is indeed optimal. These models, each featuring a different configuration, are trained using a consistent strategy and evaluated across three publicly accessible datasets.
\begin{table}[ht]\small
	\begin{center}
		\scalebox{0.9}{\begin{tabular}{c c c c c c }
				\hline
				\rule{0pt}{10pt}
				\textbf{Configuration}  & \multicolumn{2}{c}{\textbf{Metric}} & \multicolumn{1}{c}{\textbf{LOLv1}}& \multicolumn{1}{c}{\textbf{LOLv2-real}} & \multicolumn{1}{c}{\textbf{LOLv2-syn}}\\ \hline

				\multirow{3}{*}{Base}  & \multicolumn{2}{c}{PSNR}$\uparrow$
				& 19.50 & 19.03 & 21.63 \\  
				
				& \multicolumn{2}{c}{SSIM}$\uparrow$
				& 0.787 & 0.803 & 0.894 \\
				
				& \multicolumn{2}{c}{LPIPS} $\downarrow$
				& 0.1697 & 0.2132 & 0.0805 \\  
				& \multicolumn{2}{c}{MAE} $\downarrow$
				& 0.1260 & 0.1376 & 0.0891\\  
				
				\hline
				\multirow{3}{*}{+IFA}  & \multicolumn{2}{c}{PSNR} $\uparrow$ & 20.93 & \textcolor{blue}{20.45} & \textcolor{red}{24.63} \\  
				
				& \multicolumn{2}{c}{SSIM } $\uparrow$ & 0.821 & \textcolor{blue}{0.844} & \textcolor{red}{0.933} \\
				& \multicolumn{2}{c}{LPIPS}$\downarrow$
				& \textcolor{blue}{0.1068} & \textcolor{blue}{0.1061} & \textcolor{red}{0.0385} \\		
				& \multicolumn{2}{c}{MAE}$\downarrow$
				& 0.1082 & \textcolor{blue}{0.1231} & \textcolor{red}{0.0703}\\  
				\hline
				\multirow{3}{*}{+TCM}  & \multicolumn{2}{c}{PSNR} $\uparrow$ & \textcolor{blue}{22.33} & 18.71 & 23.39 \\  
				
				& \multicolumn{2}{c}{SSIM } $\uparrow$ & \textcolor{blue}{0.830} & 0.818 & 0.922\\
				& \multicolumn{2}{c}{LPIPS}$\downarrow$
				& 0.1181 & 0.1227 & 0.0532 \\	
				& \multicolumn{2}{c}{MAE}$\downarrow$
				& \textcolor{blue}{0.0877} & 0.1480 & 0.0751\\  
				\hline
				\multirow{3}{*}{+IFA+TCM}  & \multicolumn{2}{c}{PSNR} $\uparrow$
				& \textcolor{red}{24.01} & \textcolor{red}{21.12} & \textcolor{blue}{24.48}  \\  
				
				& \multicolumn{2}{c}{SSIM } $\uparrow$
				& \textcolor{red}{0.863} & \textcolor{red}{0.846} &  \textcolor{blue}{0.929}\\
				& \multicolumn{2}{c}{LPIPS}$\downarrow$
				& \textcolor{red}{0.0747} & \textcolor{red}{0.1001} & \textcolor{blue}{0.0396} \\	
				& \multicolumn{2}{c}{MAE}$\downarrow$
				& \textcolor{red}{0.0756} & \textcolor{red}{0.1173} & \textcolor{blue}{0.0718}\\  
				
				%
				%
				\hline
		\end{tabular}}
		\vspace{0.1in}
		\caption{The Detailed ablation analysis of each component under various training configurations reveals that the configuration incorporating all the designed components yields the best results.}
		\label{tab:component}
		\vspace{-0.35in}
	\end{center}
\end{table}

Based on the average and standard deviation of PSNR, SSIM, LPIPS, and MAE as presented in Table \ref{tab:component}, the performance of Base has achieved comparable scores, demonstrating its efficacy.
As illustrated in Fig. \ref{ablation_module_image}, on the basis of the Base module, integrated IFA can achieving the higher average values but this results in the loss of detail.
In comparison, TCM can make them in images cleaner, which makes the image visual better.
After combining the TCM and IFA, the result can be more vivid and natural appearance that enhances visual perception.
It not only achieves the highest indicators but also has a good visual effect. Harmonization of metric-oriented and visually friendly results is achieved.

\subsubsection{Hyperparameters Experiments}
\begin{table}[ht]\small
	\begin{center}
		\scalebox{1}{\begin{tabular}{c c c c c c c c c }
				\hline
				\rule{0pt}{10pt}
				\textbf{Blocks}  & \multicolumn{2}{c}{\textbf{PSNR}}$\uparrow$ & \multicolumn{2}{c}{\textbf{SSIM}}$\uparrow$ & \multicolumn{2}{c}{\textbf{LPIPS}}$\downarrow$ & \multicolumn{2}{c}{\textbf{MAE}}$\downarrow$\\ \hline
				
				\multirow{2}{*}{3}   & \multicolumn{2}{c}{\multirow{2}{*}{\centering 23.38}}
				& \multicolumn{2}{c}{\multirow{2}{*}{\centering 0.852}} & \multicolumn{2}{c}{\multirow{2}{*}{\centering 0.0876}} & \multicolumn{2}{c}{\multirow{2}{*}{\centering 0.0767}}
				\\
				\multirow{2}{*}{5}   & \multicolumn{2}{c}{\multirow{2}{*}{\centering 22.49}}
				& \multicolumn{2}{c}{\multirow{2}{*}{\centering 0.845}} & \multicolumn{2}{c}{\multirow{2}{*}{\centering 0.0847}}  & \multicolumn{2}{c}{\multirow{2}{*}{\centering 0.0876}}
				\\
				\multirow{2}{*}{8}   & \multicolumn{2}{c}{\multirow{2}{*}{\centering 23.16}}
				& \multicolumn{2}{c}{\multirow{2}{*}{\centering 0.854}} & \multicolumn{2}{c}{\multirow{2}{*}{\centering 0.0863}}  & \multicolumn{2}{c}{\multirow{2}{*}{\centering 0.0816}}
				\\
				\multirow{2}{*}{12}   & \multicolumn{2}{c}{\multirow{2}{*}{\centering 22.46}}
				& \multicolumn{2}{c}{\multirow{2}{*}{\centering 0.857}} & \multicolumn{2}{c}{\multirow{2}{*}{\centering 0.0788}}  & \multicolumn{2}{c}{\multirow{2}{*}{\centering 0.0862}}
				\\
				\multirow{2}{*}{\textbf{15}}   & \multicolumn{2}{c}{\multirow{2}{*}{\centering \textbf{\textcolor{red}{24.01}}}}
				& \multicolumn{2}{c}{\multirow{2}{*}{\centering \textbf{\textcolor{red}{0.863}}}}
				& \multicolumn{2}{c}{\multirow{2}{*}{\centering \textbf{\textcolor{red}{0.0747}}}}  & \multicolumn{2}{c}{\multirow{2}{*}{\centering \textbf{\textcolor{red}{0.0756}}}}
				\\
				\multirow{2}{*}{20}   & \multicolumn{2}{c}{\multirow{2}{*}{\centering 23.64}}
				& \multicolumn{2}{c}{\multirow{2}{*}{\centering 0.858}} & \multicolumn{2}{c}{\multirow{2}{*}{\centering 0.0773}}  & \multicolumn{2}{c}{\multirow{2}{*}{\centering 0.0816}}
				\\ \\

				\hline
		\end{tabular}}
		\vspace{0.1in}
		\caption{
			PSNR, SSIM, LPIPS, and MAE evaluation of various numbers of Residual Textual guide Fusion Block(RTFB) on LOLv1 dataset.}
		\label{tab:block}
		\vspace{-0.35in}
	\end{center}
\end{table}

In the experiment of hyperparameters, we mainly introduce the Residual Textual guide Fusion Block(RTFB) and loss function selected by the model.
We configure different numbers of RTFB: 3, 5, 8, 12, 15 and 20.
We train models with different numbers of RTFB using the same parameters
After comparison in the LOLv1 dataset, the number of 15 achieve the best performance(as shown in Table \ref{tab:block} and Fig. \ref{ablation_block_image}).

In evaluating different loss functions, we initially selected the basic smooth L1 loss and the recently popular SSIM loss for image processing, before ultimately settling on a combination of SSIM and L1 loss.
Experimental results, as detailed in Table \ref{tab:loss}, demonstrate that the SSIM loss effectively enhance the performance of NaLSuper for improving ssim.
The smooth L1 loss demonstrates effectively  enhance the performance of NaLSuper for improving psnr.
The combination of SSIM and L1 loss, which quantifies the differences between output and ground truth features, effectively aids the model in accurately restoring image color, texture, and fine details.

\begin{table}[ht]\small
	\begin{center}
		\scalebox{0.9}{\begin{tabular}{c c c c c c }
				\hline
				\rule{0pt}{10pt}
				\textbf{LOSS}  & \multicolumn{2}{c}{\textbf{Metric}} & \multicolumn{1}{c}{\textbf{LOLv1}}& \multicolumn{1}{c}{\textbf{LOLv2-real}} & \multicolumn{1}{c}{\textbf{LOLv2-syn}}\\ \hline
							
				\multirow{3}{*}{L1}  & \multicolumn{2}{c}{PSNR}$\uparrow$
				& \textcolor{blue}{22.84} & \textcolor{blue}{19.43} & \textcolor{blue}{23.83} \\  
				
				& \multicolumn{2}{c}{SSIM}$\uparrow$
				& 0.816 & 0.813 & 0.920 \\
				
				& \multicolumn{2}{c}{LPIPS}$\downarrow$
				& 0.1114 & 0.1406 & \textcolor{blue}{0.0509} \\
				& \multicolumn{2}{c}{MAE}$\downarrow$
				& \textcolor{blue}{0.0788} & \textcolor{blue}{0.1352} & \textcolor{blue}{0.0748} \\
				\hline
				\multirow{3}{*}{SSIM}  & \multicolumn{2}{c}{PSNR}$\uparrow$ & 20.87 & 19.29 & 23.35\\  
				
				& \multicolumn{2}{c}{SSIM }$\uparrow$ & \textcolor{blue}{0.831} & \textcolor{blue}{0.831} & \textcolor{blue}{0.921} \\
				& \multicolumn{2}{c}{LPIPS}$\downarrow$
				& 0.0952 & \textcolor{blue}{0.1114} & 0.0599 \\		
				& \multicolumn{2}{c}{MAE}$\downarrow$
				& 0.1063 & 0.1422 & 0.0762 \\
				\hline
				\multirow{3}{*}{L1+SSIM}  & \multicolumn{2}{c}{PSNR}$\uparrow$ & \textcolor{red}{24.01} & \textcolor{red}{21.12} & \textcolor{red}{24.48} \\  
				
				& \multicolumn{2}{c}{SSIM }$\uparrow$ & \textcolor{red}{0.863} & \textcolor{red}{0.846} & \textcolor{red}{0.930}\\
				& \multicolumn{2}{c}{LPIPS}$\downarrow$
				& \textcolor{red}{0.0747} & \textcolor{red}{0.1001} & \textcolor{red}{0.0396} \\	
				& \multicolumn{2}{c}{MAE}$\downarrow$
				& \textcolor{red}{0.0756} & \textcolor{red}{0.1173} & \textcolor{red}{0.0718} \\
				\hline
				
				\hline
			\end{tabular}}
		\vspace{0.1in}
		\caption{PSNR, SSIM, LPIPS, and MAE evaluation of different loss functions in LOLv1, LOLv2-real and LOLv2-syn dataset}
		\label{tab:loss}
		\vspace{-0.35in}
	\end{center}
\end{table}

\section{Conclusion}
In this paper, we introduce a Natural Language Supervision network (NaLSuper) designed to enhance images taken in low-light conditions.
This network addresses several challenges simultaneously, including low contrast, insufficient lighting, noise, and color distortions.
Firstly, we design a Textual Guidance Conditioning Mechanism (TCM) to incorporating the connections between image regions and sentence words.
In order to effectively identify and merge features from various levels of image and textual information, we design a Information Fusion Attention (IFA) module to do different levels of enhancement for different regions.
Extensive experiments shows that NaLSuper effectively handles a wide range of low-light images and significantly surpasses current methods in performance.

\bibliographystyle{plain}
\bibliography{NaLSuper-Manuscript}

\end{document}